\documentclass{article} % For LaTeX2e
\usepackage{iclr2021_conference,times}

\usepackage{amsmath,amsfonts,bm}

\def\eqref#1{equation~\ref{#1}}
\def\1{\bm{1}}

\DeclareMathAlphabet{\mathsfit}{\encodingdefault}{\sfdefault}{m}{sl}
\SetMathAlphabet{\mathsfit}{bold}{\encodingdefault}{\sfdefault}{bx}{n}

\DeclareMathOperator*{\argmax}{arg\,max}

\usepackage{hyperref}
\usepackage{url}
\usepackage[ruled,vlined]{algorithm2e}
\usepackage{enumitem}

\usepackage{adjustbox}
\usepackage{array}
\usepackage{booktabs} % professional-quality tables

\usepackage{pifont}% http://ctan.org/pkg/pifont
\newcommand{\cmark}{\ding{51}}%

\usepackage{tabularx}% http://ctan.org/pkg/tabularx
\usepackage{pgfplots}
\usepackage{tikz}
\usepgfplotslibrary{fillbetween}

\newcolumntype{R}[2]{%
    >{\adjustbox{angle=#1,lap=\width-(#2)}\bgroup}%
    l%
    <{\egroup}%
}
\newcommand*\rot{\multicolumn{1}{R{45}{1em}}}% no optional argument here, please!

\DontPrintSemicolon

\newcommand{\Le}[1]{{\color{red}{\bf\sf [Le: #1]}}}
\newcommand{\Karan}[1]{{\color{blue}{\bf\sf [Karan: #1]}}}

\newcommand{\bh}[1]{{\color{blue}{[Binghong: #1]}}}
\title{\Large How to Design Sample and Computationally Efficient VQA Models}
\newcommand{\PROJNAME}{DePe}
\author{Karan Samel \\
Georgia Tech \\
ksamel@gatech.edu \\
\And
Zelin Zhao \\
Shanghai Jiao Tong University \\
sjtuytc@sjtu.edu.cn \\
\And
Binghong Chen \\
Georgia Tech \\
binghong@gatech.edu \\
\And
Kuan Wang \\
Georgia Tech \\
kuanwang@gatech.edu \\
\And
Robin Luo \\
Georgia Tech \\
robin1997@gatech.edu \\
\And
Le Song \\
Mohamed bin Zayed University of AI \\
le.song@mbzuai.ac.ae}

\iclrfinalcopy % Uncomment for camera-ready version, but NOT for submission.
\begin{document}

\maketitle

\begin{abstract}
In multi-modal reasoning tasks, such as visual question answering (VQA), there have been many modeling and training paradigms tested. Previous models propose different methods for the vision and language tasks, but which ones perform the best while being sample and computationally efficient? Based on our experiments, we find that representing the text as probabilistic programs and images as object-level scene graphs best satisfy these desiderata. We extend existing models to leverage these soft programs and scene graphs to train on question answer pairs in an end-to-end manner. Empirical results demonstrate that this differentiable end-to-end program executor is able to maintain state-of-the-art accuracy while being sample and computationally efficient.
\end{abstract}

% \begin{abstract}
% We present a  (DePe), which addresses Visual Question Answering (VQA) in a sample and computationally efficient manner. \PROJNAME \ parses the question into probabilistic programs and softly executes them to acquire the final answer. These functional programs adopt soft-logic functions to enable approximate probabilistic logic reasoning. In addition to the language, \PROJNAME \ also jointly learns visual object-centric representations in an end-to-end manner. We demonstrate through extensive experiments that \PROJNAME \ is more sample and computationally efficient than other VQA methodologies while retaining state-of-the-art performance.
% \end{abstract}

%  We develop a differentiable function program executor which treats natural language questions as programs and differentiably execute them to compose soft-logic functions to extract answers from input images. Our model achieve sample efficiency by incorporating structures in both the differentiable function program executor and the soft-logic functions: the former component in-corporates a soft selection mechanism and a stack structure, and the latter uses soft-logic to perform approximate probabilistic logic reasoning. In our experiments, we show that our methods, called DiffVQA achieves state-of-the-art results, while being sample efficient.
\section{Introduction}

Many real-world complex tasks require both perception and reasoning (or System I and System II intelligence \citep{sutton_rl}), such as VQA. What is the best way to integrate perception and reasoning components in a single model? Furthermore, how would such an integration lead to accurate models, while being sample and computationally efficient? Such questions are important to address when scaling reasoning systems to real world use cases, where empirical computation bounds must be understood in addition to the final model performance.

% different models.
There is a spectrum of methods in the literature exploring different ways of integrating perception and reasoning. Nowadays, the perception is typically carried out via neural models: such as CNNs for vision, and LSTMs \citep{gers1999learning} or Transformers \citep{vaswani2017attention} for language. Depending on the representation of perception input and their reasoning interface, a method can be either more towards the neural end of the spectrum or more toward the symbolic end. 

% More-neural methods implicitly encode the perception and reasoning into latent parameters, but require large data sets to perform well. Symbolic approaches explicitly encode the perception and reasoning with known priors, but are less robust to imperfect data.

% There are neural pixel-level and symbolic object-level representations when considering visual perception. approaches vary within perception and reasoning methods as well. 
For the vision part, models can either use pixel-level or object-level symbolic representation. For the language part, models can generate either textual attention or programs, where the text is decomposed into a sequence of functions. Within the program representations, models typically operate on a selected discrete program or on probabilistic programs. The reasoning part used to produce the final answer can either use neural models, symbolic reasoning, or something in between, such as neural module networks (NMN) or soft logic blocks. 

Existing works for NMN methods leverage pixel-level representations and program representations such as NMN \citep{LearningToReason}, Prob-NMN \citep{vedantam2019probabilistic}, and Stack-NMN \citep{stack_nmn}. Representative models that use object-level vision also leverage both neural and symbolic language and reasoning. Models that are more neural are LXMERT \citep{lxmert} and NSM \citep{NSM}, while those that are more symbolic are NS-VQA \citep{yi2018neural}, NS-CL \citep{NSCL} and NGS \citep{NGS}. A systematic comparison across these models is illustrated in Table \ref{tab.compare} with more details in Appendix \ref{apx.compare}. 

Overall, neural models have more expressive power but with more parameters, while more-symbolic models have more prior structures built into them but with fewer parameters. There is an interesting bias-variance trade-off in the model design. By encoding as much bias into the model as possible, one could reduce sample requirements.   

The different choices of perception and reasoning components also limit how the QA models will be trained. If both components are chosen as neural modules, then the training can be done in a very efficient end-to-end fashion. If the reasoning is carried out using more discrete operations, then the perception model needs to sample discrete outputs or take discrete inputs to interface with downstream reasoning. For instance, if symbolic reasoning is used, REINFORCE \citep{williams1992simple} is typically used to train the perception models, which may require many samples during the optimization process. Alternatively, one can also use expensive abduction \citep{NGS} to manipulate the perception models outputs to provide the correct reasoning and then optimize these perception models using these pseudo-labels. Overall, more neural models will be easier to optimize, while more symbolic models will need additional expensive discrete sampling during optimization. To highlight this interesting fact, we call it the neuro-symbolic trade-off.

This neuro-symbolic trade-off also affects sample efficiency and computational efficiency. To be more sample efficient, the model needs to be less neural, yet, a more neural model can be more computationally efficient during training. Thus a method that can achieve an overall good performance in terms of both sample and computation efficiency will require systematically determining which perception and reasoning components should be used and how to integrate them. To design such a model, we first test which method within each perception and reasoning component works the most efficiently. From this neuro-symbolic trade-off exploration we can design a model that uses these most efficient components and compare its overall performance against existing models.

\begin{table}
\renewcommand*\rot{\multicolumn{1}{R{50}{1em}}}% no optional argument here, please!
% \begin{table}[h]
\setlength{\tabcolsep}{10pt}
\begin{tabular}{r|cc|ccc|ccc|ccc}
\toprule
& \multicolumn{2}{c}{Vision}
& \multicolumn{3}{|c}{Language}
& \multicolumn{3}{|c}{Reasoning}
& \multicolumn{3}{|c}{Training} \\
\midrule 
& 
\rot{Pixel Att.} &
\rot{Object-level} &
% \rot{Graph} &
\rot{Text Att.} &
\rot{Symbolic Progs.} &
\rot{Soft Progs.} &
\rot{Neural} &
\rot{Symbolic} &
\rot{Soft Logic} &
\rot{End-to-End} &
\rot{Sampling} \\
\midrule
LXMERT & & \cmark & \cmark & & & \cmark & & & \cmark &\\ 
(Prob-)NMN & \cmark & & & \cmark & & & & \cmark & \cmark & \cmark \\ 
Stack-NMN & \cmark & & & & \cmark & & & \cmark & \cmark & \\ 
NSM & & \cmark & \cmark & & & \cmark & & & \cmark & \\ 
NS-VQA & & \cmark & & \cmark & & & \cmark & & \cmark & \cmark\\
NS-CL & & \cmark & & \cmark & & & & \cmark & \cmark & \cmark\\ 
NGS & & \cmark & & \cmark & & & \cmark & & & \cmark \\ 
% \midrule 
% Ours & & \cmark & & & \cmark & & & \cmark & \cmark & & \\
\bottomrule
\end{tabular}
\caption{A breakdown of VQA models by indicating which method is used with respect to their vision, language, inference, and training components. Refer to Appendix \ref{apx.compare} for a detailed description of these methods.}
\label{tab.compare}
\end{table}

% \begin{table}
% \renewcommand*\rot{\multicolumn{1}{R{50}{1em}}}% no optional argument here, please!
% % \begin{table}[h]
% \setlength{\tabcolsep}{10pt}
% \begin{tabular}{l|l|l|l|l}
% \toprule
%          & Vision          & Language         & Reasoning  & Training   \\ \midrule
% LXMERT \citep{lxmert}   & Object & Attention & Neural     & E2E \\ 
% NMN \citep{LearningToReason}     & Attention & Symbolic         & Soft Logic & E2E \\
% StackNMN \citep{stack_nmn}& Attention & Attention & Soft Logic & E2E \\ 
% NSM \citep{NSM}     & Object & Attention & Neural     & E2E \\ 
% NSCL \citep{NSCL}    & Object & Symbolic         & Symbolic   & RL  \\ 
% NGS \citep{NGS}     & Object & Symbolic         & Symbolic   & RL  \\ \midrule
% Ours     & Object & Attention & Soft Logic & E2E \\ \bottomrule
% \end{tabular}
% \caption{Comparison with other models w.r.t. vision, language, and training. We use E2E, RL to denote end-to-end training and reinforcement learning (REINFORCE) accordingly.}
% \end{table}

\section{Problem Setting}
Before the exploration, we formally define the different choices for the vision, language, and reasoning components. In the general VQA setting we are provided with an image $I$, a natural language question $Q$, and an answer $A$. We now define how these basic inputs are used in each component.

\subsection{Representation for Vision}
Given the image $I$ there are two predominant visual representations: pixel and object-level attention.

{\bf Pixel Attention.} Given an image  one can leverage traditional deep learning architectures used for image representations and classification such as ResNets \citep{he2016deep}. Here the image is passed through many residual convolution layers before entering a MLP sub-network to perform a classification task. From one of these MLP linear layers, an intermediate dense image representation feature $f_I \in \mathbb{R}^D$ can be extracted, denoted by $f_I = \text{ResNet}(I)$. These features are used further down the VQA pipeline, where the downstream model computes attention over the relevant part of the feature based on the question asked.

{\bf Object-level.} Another paradigm is to leverage object detection models such as Faster R-CNNs \citep{ren2015faster} to identify individual objects within images. Given objects in the image, one can conduct more object-level or symbolic reasoning over the image, instead of reasoning through a pixel by pixel basis. 

In this object-level representation, a set of object location bounding boxes (bbox) can be detected and labeled directly by using R-CNN as $O = \{(\text{bbox}_1, \text{label}_1), \cdots, (\text{bbox}_T, \text{label}_T)\} = \text{R-CNN}(I)$ for a preset number of $T$ objects. Here $o \in O$ can be labeled as ``small red shiny ball'' or ``large green tray'' based on what is in the image. 

Another approach is to factor the joint bounding box and label prediction into individual components to be handled by separate models. First the bounding boxes are extracted from the R-CNN as $\{\text{bbox}_i\}_{i=1}^{T} = \text{R-CNN}(I)$. Then these can be passed into a separate MLP network to retrieve the labels $\{\text{label}_i\}_{i=1}^{T} = \text{MLP}(\text{ResNet}(I[\text{bbox}_i]))$, where $I[\text{bbox}]$ is cropped image at that bounding box location. These can be used to define the final set of objects: $O = \{(\text{bbox}_i, \text{label}_i)_{i=1}^{T}\}$. 

In such a setup, the benefit is that the R-CNN can be trained just as an object detector for a generic object class versus the background, whose annotations are easier to obtain. Furthermore, the number of supervised data the label MLP uses for training can be controlled separately. This is a useful mechanic during our model efficiency analysis where we work under the assumption that object bounding box is almost perfect, while object labeling is imperfect and expensive to annotate.

\subsection{Representation for Language}

The language representations operates on the natural text question $Q$. Some data sets also provide intermediate representations of each $Q$ through a function program layout $FP$. $FP$ represents the question as a sequence of abstract functions $\mathcal{F}$ as $FP = [\mathcal{F}_1, \mathcal{F}_2, ..., \mathcal{F}_t]$ for $\mathcal{F}_i \in \mathcal{F}$. These function programs are used jointly with the visual representations to conduct reasoning to arrive at answer $A$. Details about potential realizations of $\mathcal{F}$ are described in the following reasoning representation section \ref{subsection:reasoning}. Given the question $Q$ and its representation $FP$ we can define different approaches for representing the text.

{\bf Text Attention.} Just using the embedded text tokens $E$ a model can embed the question $Q$ through a recurrent network to generate a final question representation $h_T$, where $T$ is the maximum length sequence. Then $h_T$ can be put through an recurrent decoder to obtain a latent function at each step $c_t$ through an attentive combination of the hidden states $c_t = \sum_T a_t \cdot h_t$.

{\bf Symbolic Program.} If we want to explicitly produce a $FP$ for a corresponding question $Q$, we similarly encode the text as done for text attention. During decoding $c_t$ is passed through a MLP to predict a valid function token. Then the most likely program is selected as $\argmax_{FP} P(FP \mid Q)$.

{\bf Soft Program.} When choosing a discrete symbolic program, the uncertainty of other function program parses is thrown out. Instead the probabilities for each program can be saved and an expected program can be computed as $\mathbb{E}[P(FP \mid Q)]$. Intuitively all the possible programs have to be considered in this scenario which can be intractable. Instead soft program methods such as Stack-NMN factor this as $\mathbb{E}[P(FP \mid Q)] = \mathbb{E}[\prod_T P(\mathcal{F}_t \mid Q)] = \prod_T \mathbb{E}[P(\mathcal{F}_t \mid Q)]$. This enables preserving a distribution of functions at each step instead of selecting a single one.

\subsection{Representation for Reasoning}
\label{subsection:reasoning}

Given the visual and language representations, the reasoning component use these representations to derive the final answer $A$. Here we discuss methods that are neural, symbolic, and soft logic based.

{\bf Neural Reasoning.} 
Reasoning can be made directly with the image feature $f_I$ and encoded question $h_T$ such as $A = \text{MLP}([h_T; f_I])$ in a purely neural fashion. Other approaches can leverage the object representations $O$. This is done by modulating the attention over which $O$ correspond to final answer $A$, conditioned on $h_T$, as done in NSM or LCGN. LXMERT uses cross-modal attention between text embeddings $E$ and $O$ to predict the final answer. All these methods are more neural, but the $FP$ can be leveraged as well to incorporate better biases through symbolic and soft programs.

% \Zelin{This sentence is ambiguous. What do you mean by "directly with image"? And what is "not directly with image"? This example is rather confusing, are you referring to a baseline that passes through an image to a MLP, without even encoding the question?}

{\bf Symbolic Representations.} From the question we can define abstract functions $\mathcal{F}$ to generate $FP$ as described in the previous section. Representing $\mathcal{F}$ in a symbolic form enables encoding general knowledge or certain dataset's domain specific language (DSL) into a model. This improves model interpretability and provides better inductive biases as well. Here we further describe two classes of these functions: fine grained and coarse.  

% \Zelin{We need to clearly define what symbolic representations are.}

% \Zelin{Why do you mention bias encoding here? Isn't this paragraph introducing symbolic representations?} <- We want to motivate the use of such structures.

A fine grained representation of $FP$ is sequence of n-ary predicates, functions with n arguments, composing $\mathcal{F}$. For example, given the question $Q = \text{``What is the shape of the thing left of the sphere?''}$, a sample fine grained program can be defined as $FP = [\texttt{filter\_shape(sphere, O)}, \texttt{relate(left, O)}, \texttt{query\_shape(O)}]$ Here the visual representation ($O$ or $f_I$) and higher level argument concepts, such as \texttt{sphere}, are used as inputs to each function.  We observe clear biases encoded into the function architecture, as given a scene graph of objects $O$ and their relations, one could walk along this graph using $FP$ to get the final answer $A$. The trade-off is that the model has to deal with more complex functions, whose input arguments and output types can vary. For example $\texttt{relate\_shape}$ and $\texttt{relate}$ return a subset of objects, while $\texttt{query\_shape}$ returns a string. Such formulations are used by more neuro-symbolic methods such as NS-VQA and NS-CL.

Coarse function types consist of simpler predicates whose arity is fixed, typically 1, over $\mathcal{F}$. Given the previous question $Q$, a coarse function can be defined as  $FP = [\texttt{filter}_\theta(f_{I}), \texttt{relate}_\theta(f_{I}), \texttt{query}_\theta(f_{I})]$. Here less structure is required with respect to the language and visual representation where each function can be parameterized as a NMNs. These require more parameters than DSL functions but are syntactically easier to handle as they typically just operate on a fixed dimensional image feature $f_I$, thus implicitly encoding the function arguments.

{\bf Symbolic Reasoning.} Using any coarse or fine representation type for $\mathcal{F}$, the symbolic reasoning can take place over the selected symbolic program $FP$. We define the high level execution of the symbolic reasoner to arrive at the answer by executing over $FP$ as $A = \langle FP, \text{image representation} \rangle_S$. In the fine grained and coarse samples this would look like: 

\[A = \langle FP, O \rangle_S = \texttt{query\_shape(relate(left, filter\_shape(sphere, O)))} \]
\[A = \langle FP, f_I \rangle_S = \texttt{query}_{\theta}\texttt{(relate}_{\theta}\texttt{(filter}_{\theta}\texttt{(}f_I\texttt{)))} \]

Since the structure of the reasoning is discrete, to update the visual and language model weights requires sampling based learning such as REINFORCE or abductive methods.

{\bf Soft Logic Reasoning.} When conducting the symbolic reasoning we discard the uncertainty of the visual representations when generating the labels for $O$. Instead the probabilities for $O$ can be saved. Here the uncertainty from the detections can be propagated during the execution in a soft manner to arrive at $A$. We can similarly define the soft logic reasoning as $A = \langle FP, I \rangle_{SL} = \mathbb{E}_{O \sim \text{R-CNN}(I)}[\langle FP, O \rangle_S]$. Due to the probabilistic execution, this can be optimized end-to-end. 

% \Karan{from http://cs.brown.edu/people/sbach/files/kimmig-probprog12.pdf, the more formal definitions, "relaxation of the boolean logic using the luk t-norm, a more formal statement}

Now that the components and their corresponding methods have been defined, we explore which methods are the most efficient for their respective task.

\section{Neuro-Symbolic Trade-Off Exploration}

Many deep VQA models have been developed in the literature with a range of design elements, as can be seen from Table \ref{tab.compare}. Which one of these methods is the key factor in making a deep VQA model sample efficient while at the same time achieving state-of-the-art? In this comparison using the CLEVR dataset \citep{johnson2017clevr}, we aim to understand which design elements individually perform the best. Based on these findings, better end-to-end models can be designed from the individual methods selected. More specifically, we will explore the sample and computational efficiency for the following components: 
\begin{itemize}
    \item Visual representations through pixel attention and object-level methods. 
    \item Reasoning execution through neural modules, symbolic execution, and soft logic. 
    \item Language encoding for questions through text attention, symbolic and soft programs. 
\end{itemize}

For the representations of language and reasoning, we observe that these two components are tightly coupled. In language we define $\mathcal{F}$ and the corresponding $FP$ given $Q$. In reasoning these functions get executed for fine grained functions, or a network gets constructed from neural modules in the coarse case. For this reason we found it difficult to isolate the language and reasoning exploration. This motivated us to initially observe the interactions between the vision and reasoning given a fixed $FP$. Then by iteratively selecting the best vision and reasoning components, we explore the most efficient language and reasoning combination. Each method is also explained in more detail in Appendix \ref{apx.exp}.

\subsection{Visual Reasoning}
To test the visual perception and the reasoning components we break down the tests into two parts. First we first determine which visual representation is more efficient: pixel attention or object-centric. Second we find the reasoning method that best complements the selected visual representation.

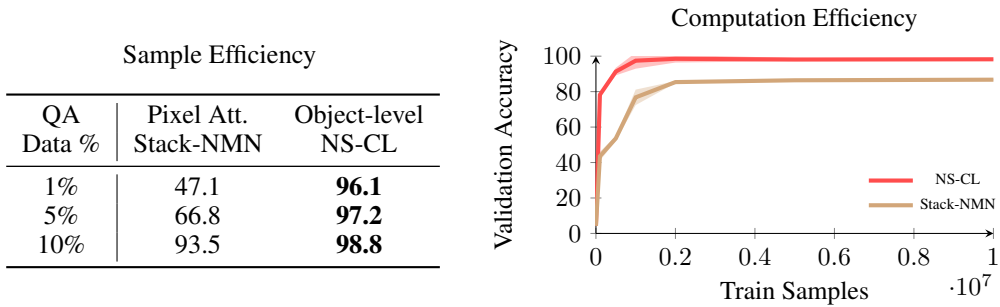
\begin{figure}[!h]
    % \caption{Global caption}
    \begin{minipage}{.49\linewidth}
        % \renewcommand{\arraystretch}{1.28}
        % \captionof{Sample efficiency.}
        \centering
        Sample Efficiency
        \vspace{3mm}
        
        \begin{tabular}{c| c c}
        \toprule
        QA & Pixel Att. & Object-level \\
        Data \% & Stack-NMN & NS-CL \\
        \midrule
        1\% & 47.1 & \textbf{96.1} \\
        5\% & 66.8 & \textbf{97.2} \\
        10\% & 93.5 & \textbf{98.8} \\
        \bottomrule
        \end{tabular}
    \end{minipage}
    \begin{minipage}{.51\linewidth}
        \resizebox{1.0\textwidth}{!}{%
        \begin{tikzpicture}
\begin{axis}[
    % xmode=log
    % xtick={0, 120, 600, 1200, 6000},
    title={Computation Efficiency},
    ytick distance=20,
    ymin=0,
    ymax=100,
    height=4cm,
    width=7cm,
    axis lines = left,
    xlabel = Train Samples,
    ylabel = Validation Accuracy,
    no markers,
    every axis plot/.append style={ultra thick},
    legend style={at={(0.85,0.4)},anchor=north, legend columns=1, font=\tiny, draw=none}
]

\addplot [color=red!70, mark=square] table[x=samples,y=nscl] {data/stat_efficiency_vision_scratch.dat};
\addlegendentry{NS-CL}

\addplot [color=brown!70, mark=square] table[x=samples,y=snmn] {data/stat_efficiency_vision_scratch.dat};
\addlegendentry{Stack-NMN}

% \addplot [color=blue!70, mark=square] table[x=samples,y=nsvqa] {data/stat_efficiency_vision_scratch.dat};
% \addlegendentry{NS-VQA}

% std
\addplot [name path=upper,draw=none] table[x=samples,y expr=\thisrow{nscl}+\thisrow{nscl_std}] {data/stat_efficiency_vision_scratch.dat};
\addplot [name path=lower,draw=none] table[x=samples,y expr=\thisrow{nscl}-\thisrow{nscl_std}] {data/stat_efficiency_vision_scratch.dat};
\addplot [fill=red!30, fill opacity=0.8] fill between[of=upper and lower];

\addplot [name path=upper,draw=none] table[x=samples,y expr=\thisrow{snmn}+\thisrow{snmn_std}] {data/stat_efficiency_vision_scratch.dat};
\addplot [name path=lower,draw=none] table[x=samples,y expr=\thisrow{snmn}-\thisrow{snmn_std}] {data/stat_efficiency_vision_scratch.dat};
\addplot [fill=brown!30, fill opacity=0.8] fill between[of=upper and lower];

% \addplot [name path=upper,draw=none] table[x=samples,y expr=\thisrow{ours}+\thisrow{ours_std}] {data/stat_efficiency_vision_scratch.dat};
% \addplot [name path=lower,draw=none] table[x=samples,y expr=\thisrow{ours}-\thisrow{ours_std}] {data/stat_efficiency_vision_scratch.dat};
% \addplot [fill=blue!15, fill opacity=0.8] fill between[of=upper and lower];

\end{axis}
\end{tikzpicture}
        }
    \end{minipage}%
    \caption{Sample and computational efficiency for pixel attention and object-level vision representations. The sample efficiencies experiments just use a total percentage of the available QA pairs and report the validation score after convergence. The computational efficiency is evaluated on the 10 \% QA data and are run 5 times. The train samples indicate the number of examples seen by the model over time (iteration $\times$ batch size).}
    \label{tab:vision_comp}
\end{figure}

\textbf{Pixel Attention versus Object-level.} We compare pixel attention Stack-NMN and object-level NS-CL representation methods. We chose these two models as their visual representations differ but had continuous end-to-end reasoning methods given a fixed $FP$.

We set up this experiment by training and freezing each model's language components on CLEVR question program pairs $(Q, FP)$ to isolate the effects of visual methods. Then the visual representation is trained from scratch using different percentages of CLEVR text question and answer ($Q, A$) pair data (QA Data \%), where no object level supervision is provided.

The sample and computational efficiencies are illustrated in Table \ref{tab:vision_comp}, which indicate \textbf{object-level} representations work better. Object-level detectors are able to leverage cheap object locations labels for localization and can leverage MLPs to classify objects given a dense feature $f_I \in \mathbb{R}^D$. In contrast, pixel-wise methods needs to learn the object attention maps through more parameter intensive convolutional layers over $I \in \mathbb{R}^{L \times W  \times 3}$ where $D < L \times W  \times 3$ channels.

\begin{figure}[!h]
    % \caption{Global caption}
    \begin{minipage}{.49\linewidth}
        \centering
        Sample Efficiency
        \vspace{3mm}
        
        \begin{tabular}{c| c c c}
        \toprule
        QA & & & \\
        Data \% & NS-RL & NS-AB & NS-CL \\
        \midrule
        1\% & 47.1 & 59.8 & \textbf{96.1} \\
        5\% & 72.8 & 76.6 & \textbf{97.2} \\
        10\% & 81.3 & 84.2 & \textbf{98.8} \\
        \bottomrule
        \end{tabular}
        % \caption{Sample efficiency.}
    \end{minipage}
    \begin{minipage}{.51\linewidth}
        \resizebox{1.0\textwidth}{!}{%
        \begin{tikzpicture}
\begin{axis}[
    % xmode=log
    % xtick={0, 120, 600, 1200, 6000},
    title={Computation Efficiency},
    ytick distance=20,
    ymin=0,
    ymax=100,
    height=4cm,
    width=7cm,
    axis lines = left,
    xlabel = Train Samples,
    ylabel = Validation Accuracy,
    no markers,
    every axis plot/.append style={ultra thick},
    legend style={at={(0.85,0.5)},anchor=north, legend columns=1, font=\tiny, draw=none}
]

\addplot [color=red!70, mark=square] table[x=samples,y=nscl] {data/stat_efficiency_vision_scratch.dat};
\addlegendentry{NS-CL}

\addplot [color=cyan!70, mark=square] table[x=samples,y=abduction] {data/stat_efficiency_vision.dat};
\addlegendentry{NS-AB}

\addplot [color=purple!70, mark=square] table[x=samples,y=nsrl] {data/stat_efficiency_vision.dat};
\addlegendentry{NS-RL}

% std
\addplot [name path=upper,draw=none] table[x=samples,y expr=\thisrow{nscl}+\thisrow{nscl_std}] {data/stat_efficiency_vision_scratch.dat};
\addplot [name path=lower,draw=none] table[x=samples,y expr=\thisrow{nscl}-\thisrow{nscl_std}] {data/stat_efficiency_vision_scratch.dat};
\addplot [fill=red!30, fill opacity=0.8] fill between[of=upper and lower];

\addplot [name path=upper,draw=none] table[x=samples,y expr=\thisrow{abduction}+\thisrow{abduction_std}] {data/stat_efficiency_vision.dat};
\addplot [name path=lower,draw=none] table[x=samples,y expr=\thisrow{abduction}-\thisrow{abduction_std}] {data/stat_efficiency_vision.dat};
\addplot [fill=cyan!30, fill opacity=0.8] fill between[of=upper and lower];

\addplot [name path=upper,draw=none] table[x=samples,y expr=\thisrow{nsrl}+\thisrow{nsrl_std}] {data/stat_efficiency_vision.dat};
\addplot [name path=lower,draw=none] table[x=samples,y expr=\thisrow{nsrl}-\thisrow{nsrl_std}] {data/stat_efficiency_vision.dat};
\addplot [fill=purple!30, fill opacity=0.8] fill between[of=upper and lower];

\end{axis}
\end{tikzpicture}
        }
    \end{minipage}%
    \caption{Sample and computational efficiency for soft versus symbolic reasoning. We take soft logic NS-CL and modify the reasoning to become symbolic. Then we optimize these symbolic methods through REINFORCE and abduction for NS-RL and NS-AB respectively.}
    \label{tab:vision_reason}
\end{figure}
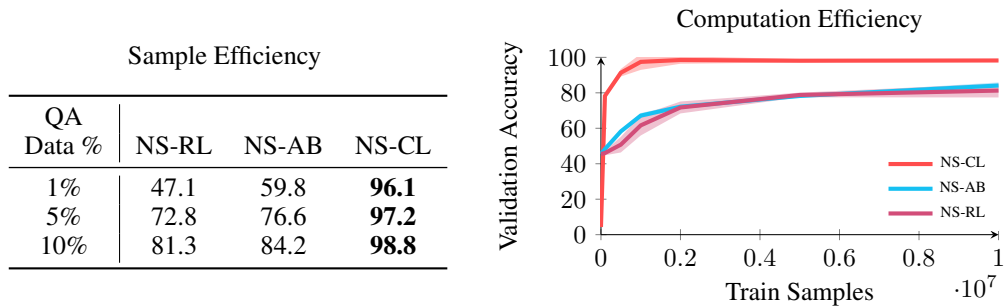

\textbf{Symbolic versus Soft Logic Reasoning.} Since we will use fine grained object-level $O$ information, we don't need to conduct parameter intensive pure neural reasoning over $f_I$. This lets us leverage of function programs $FP$. Similarly, given $O$, we don't use coarse NMN functions, which are only compatible operating over $f_I$. We focus on testing fine grained $\mathcal{F}$ to determine the best way to reason over the answer $A = \langle FP, I \rangle_{*}$, either using symbolic or soft logic.

We use NS-CL which already performs soft logic over ($Q, A$) pairs. To test the symbolic reasoning we replace the soft functions $\mathcal{F}$ defined in NS-CL by discrete counterparts, similar to the symbolic execution performed by NS-VQA. To train the symbolic reasoning with QA data, we test REINFORCE and abduction based optimization, denoted as NS-RL and NS-AB respectively. The results in Table \ref{tab:vision_reason} indicate that propagating the uncertainty of the object predictions through \textbf{soft logic} leads to gains in computational efficiency as well as final performance. Abduction has benefits over REINFORCE as it is able to selectively sample the object probabilities, which shows to be a more efficient procedure as the accuracy of the vision model increases. 

% Since all methods are trained in the same fashion, and the only difference is the reasoning during inference, we only observe the sample efficiencies, since the computational efficiencies would be the same.

At this point we have been testing the visual components and present the benefits of object-level representation and soft logic reasoning. Given these two methods, we now explore which language representation would be the most efficient to use.

\subsection{Linguistic Reasoning}

To test our language representation we want to determine the most efficient representation for $Q$. Taking into account the vision experiments, we find NS-CL's operation over object-level representations and executing soft logic reasoning over a fixed $FP$ to be the most suitable approach. To build off of this, we want to understand the best representation of $FP$ for reasoning. Therefore we look at symbolic and soft approaches. Recall the tight integration between the language representation and reasoning means that we investigate these two components in a joint fashion.

\begin{table}[!h]
    % \caption{Global caption}
    \begin{minipage}{.49\linewidth}
        \centering
        Sample Efficiency
        \vspace{3mm}
        
        \begin{tabular}{c| c c}
        \toprule
        QA & Symbolic Progs. & Soft Progs. \\
        Data \% & NS-CL & Stack-NMN\\
        \midrule
        1\% & \textbf{86.5} &  81.3\\
        5\% & 89.2 &  \textbf{90.2}\\
        10\% & 92.1 &  \textbf{94.3}\\
        \bottomrule
        \end{tabular}
        % \caption{Sample efficiency.}
    \end{minipage}
    \begin{minipage}{.51\linewidth}
        \resizebox{1.0\textwidth}{!}{%
        \begin{tikzpicture}
\begin{axis}[
    % xmode=log
    % xtick={0, 120, 600, 1200, 6000},
    title={Computation Efficiency},
    ytick distance=20,
    ymin=0,
    ymax=100,
    height=4cm,
    width=7cm,
    legend style={at={(1.0,0.0)},anchor=south east},
    axis lines = left,
    xlabel = Train Samples,
    ylabel = Validation Accuracy,
    no markers,
    every axis plot/.append style={ultra thick},
    legend style={at={(0.85,0.4)},anchor=north, legend columns=1, draw=none, font=\tiny}
]
% means
\addplot [color=red!70, mark=square] table[x=samples,y=nscl] {data/stat_efficiency_text.dat};
\addlegendentry{NS-CL}

\addplot [color=brown!70, mark=square] table[x=samples,y=snmn] {data/stat_efficiency_text.dat};
\addlegendentry{Stack-NMN}

% std
\addplot [name path=upper,draw=none] table[x=samples,y expr=\thisrow{nscl}+\thisrow{nscl_std}] {data/stat_efficiency_text.dat};
\addplot [name path=lower,draw=none] table[x=samples,y expr=\thisrow{nscl}-\thisrow{nscl_std}] {data/stat_efficiency_text.dat};
\addplot [fill=red!30, fill opacity=0.8] fill between[of=upper and lower];

\addplot [name path=upper,draw=none] table[x=samples,y expr=\thisrow{snmn}+\thisrow{snmn_std}] {data/stat_efficiency_text.dat};
\addplot [name path=lower,draw=none] table[x=samples,y expr=\thisrow{snmn}-\thisrow{snmn_std}] {data/stat_efficiency_text.dat};
\addplot [fill=brown!30, fill opacity=0.8] fill between[of=upper and lower];

\end{axis}
\end{tikzpicture}
        }
    \end{minipage}%
    \caption{Sample and computational efficiency for pixel attention and object-centric vision representations.}
    \label{tab:text_comp}
\end{table}

\textbf{Symbolic versus Soft Execution.} To test the language representation, we similarly train and freeze the visual representations to isolate the effects of the language models. These language models are then trained are also trained end-to-end on ($Q, A$) pairs without ground truth $FP$. 

We compare NS-CL, which uses fine grained symbolic programs and soft logic reasoning over $O$ to Stack-NMN which uses coarse soft programs and neural reasoning over $f_I$. For Stack-NMN the language and vision are trained jointly, but for NS-CL the language parser is trained disjointly from the vision. During testing the computation efficiency results, we equally divided the iterations by the curriculum learning setup described in their work. This led to more stable accuracy improvements over training on random samples as done in NS-VQA. 

The results are presented in Table \ref{tab:text_comp} and we generally see slower convergence than the vision trials. This is due to the fact that the program produced has to be perfect to reason correctly, while some mis-labeled objects may not lead to incorrect reasoning. Furthermore we encounter spurious predictions of incorrect $FP$ leading to the correct answer, prolonging training. 

Looking closer at the results, at 1\% QA data, the symbolic representation is on top. We posit that this is due to model exhaustively sampling the program space over an end-to-end approach given limited data. However, as the amount of data increases, the \textbf{end-to-end soft programs} show better accuracies and computational efficiency. With the results from these language and vision experiments, we now have an understanding of which methods are the most efficient for VQA.

\section{Vision and Language End-to-End Reasoning.} 

We iteratively experimented on different visual, language, and reasoning VQA components to determine which methods had the best accuracy and efficiency trade-offs. Based on this, we determined the following desiderata for efficient VQA methods:
\begin{itemize}
    \item \textbf{Soft programs} to propagate language parsing uncertainty.
    \item \textbf{Object-level} detection models with pre-trained localization for object based reasoning.
    \item \textbf{Soft logic} functions to propagate the uncertainties from the vision and language branches.
    \item \textbf{End-to-end} differentiability to efficiently optimize the perception and reasoning jointly.
\end{itemize}

We are motivated to combine the existing Stack-NMN and NS-CL frameworks we tested to synthesize such a model. However, during our trade-off exploration we found it non-trivial to simply combine different methods across visual, language and reasoning components. We address two challenges regarding storing the intermediate computation in memory and the overall optimization when using these selected methods.

\subsection{Memory}

The first challenge is incompatibility at the reasoning level. NS-CL leverages fine grained functions that leverage objects $O$. The outputs of all of Stack-NMN's soft programs that operate only on $f_I$ passed through coarse NMNs. To make such soft programs from Stack-NMN compatible with fine grained functions and object-level data, the memory storage handling the intermediate functional computation has to store object-level data as well. 

We design such a memory structure by pre-assigning different object-level data modalities to different parts of a memory matrix $M \in \mathbb{R}^{T \times (A + C + 1)}$. Here the first dimension $T$ is the stack dimension used to store intermediate function outputs for future use, similar to Stack-NMN. The second $(A + C + 1)$ dimension are the rows that store the heterogeneous values while Stack-NMN only stores a $D$ dimensional image feature $f_I$. The first $A$ dimensions in the row indicate the object set attentions $m_{det}$, or which objects the model is paying attention to at that step. The $C$ stores concatenated categorical value outputs possible by our vision models such as $m_{color}$ and $m_{size}$. The final dimension $m_{num}$ is reserved for some accumulated score such as $\texttt{count}$ or a boolean bit for true or false questions. For the CLEVR specific case the object attentions $m_{det} \in \mathbb{R}^A$. The categorical values are $[m_{color}; m_{shape}; m_{texture}; m_{size}] \in \mathbb{R}^C$, and the numeric slot $m_{num} \in \mathbb{R}$. 

This enables computing the reasoning softly over NS-CL like object-level predictions and Stack-NMN function program layouts as $\mathbb{E}_{O \sim \text{R-CNN}(I)}[\langle \prod_T \mathbb{E}[P(\mathcal{F}_t \mid Q)], O \rangle_S]$. After reasoning, the answers can be directly read from the memory instead of being inferred as a multi-class label. This is done by directly predicting the the question type from $Q$ and accessing the corresponding memory location. For example if the question was asking about the color then we would return $\argmax m_{color}$. Now that we have this fully differentiable architecture, we focus on the optimization procedure.

\subsection{Optimization}

The second challenge is that Stack-NMN trains end-to-end while NS-CL iterates between REINFORCE for program parsing and end-to-end training for visual concept learning. NS-CL fixes the language model while end-to-end training vision, while we want to jointly optimizing both language and vision models. This results in a much larger optimization landscape prone to spurious predictions from end-to-end training on QA data.

To mitigate this we initially start by training the the vision MLP and language LSTM models with a small amount of the direct supervision, using between 0.1\% - 1\% of such data. This is done over the object ($I[\text{bbox}]$, label) and language ($Q, FP$) pairs available in CLEVR. Then we train end-to-end on $(Q, A)$ pair data where we have losses for categorical and numerical answers using, cross entropy and MSE losses respectively. Additionally, we found it useful to add the initial direct supervision losses and data when training end-to-end as well, weighted by hyperparameters $\alpha$ and $\beta$. We formulate this as a regularization method that prevents the models from diverging to spurious predictions when only provided with QA labels. We further provide details and demonstrate the efficacy of this regularization in Appendix \ref{apx.optim}. All these terms give our overall loss as: 

% \Zelin{write in a probabilistic manner e.g. $P(a|\theta)$.}
\[\mathcal{L}_{E2E} = \mathcal{L}_{QA\_XEnt} + \mathcal{L}_{QA\_MSE} + \alpha \mathcal{L}_{O\_XEnt} + \beta \mathcal{L}_{FP\_XEnt}\]

From these extensions over Stack-NMN and NS-CL we construct a fully \underline{D}ifferentiable \underline{e}nd-to-end \underline{P}rogram \underline{e}xecutor (\PROJNAME). A more detailed description \PROJNAME's architecture and examples can be found in Appendix \ref{apx:depe}.

\section{Experiments}
We built \PROJNAME{} using the desiderata over efficient VQA methods and now we test its overall end-to-end performance jointly on vision and language. First we compare it to our base NS-CL and Stack-NMN methods in terms of our desired sample and computational complexity. Then we compare \PROJNAME's accuracy against other VQA methods.

\subsection{Efficiency Performance}
\begin{figure}[!h]
    % \caption{Global caption}
    \begin{minipage}{.49\linewidth}
        \centering
        Sample Efficiency
        \vspace{3mm}
        
        \begin{tabular}{c| c c c}
        \toprule
        QA & & & \\
        Data \% & Stack-NMN & NS-CL & \PROJNAME \\
        \midrule
        1\% & 43.6 & \textbf{68.1} & 65.4\\
        5\% & 66.6 & 86.7 & \textbf{87.7}\\
        10\% & 80.6 & \textbf{98.8} & 98.0\\
        \bottomrule
        \end{tabular}
        
        % \caption{Sample efficiency.}
    \end{minipage}
    \begin{minipage}{.51\linewidth}
        \resizebox{1.0\textwidth}{!}{%
        \begin{tikzpicture}
\begin{axis}[
    % xmode=log
    % xtick={0, 120, 600, 1200, 6000},
    title={Computation Efficiency},
    ytick distance=20,
    ymin=0,
    ymax=100,
    height=4cm,
    width=7cm,
    axis lines = left,
    xlabel = Train Samples,
    ylabel = Validation Accuracy,
    no markers,
    every axis plot/.append style={ultra thick},
    legend style={at={(0.85,0.5)},anchor=north, legend columns=1, font=\tiny, draw=none}
]

\addplot [color=red!70, mark=square] table[x=samples,y=nscl] {data/e2e_efficiency.dat};
\addlegendentry{NS-CL}

\addplot [color=brown!70, mark=square] table[x=samples,y=snmn] {data/e2e_efficiency.dat};
\addlegendentry{Stack-NMN}

\addplot [color=blue!70, mark=square] table[x=samples,y=ours] {data/e2e_efficiency.dat};
\addlegendentry{\PROJNAME}

% std
\addplot [name path=upper,draw=none] table[x=samples,y expr=\thisrow{nscl}+\thisrow{nscl_std}] {data/e2e_efficiency.dat};
\addplot [name path=lower,draw=none] table[x=samples,y expr=\thisrow{nscl}-\thisrow{nscl_std}] {data/e2e_efficiency.dat};
\addplot [fill=red!30, fill opacity=0.8] fill between[of=upper and lower];

\addplot [name path=upper,draw=none] table[x=samples,y expr=\thisrow{snmn}+\thisrow{snmn_std}] {data/e2e_efficiency.dat};
\addplot [name path=lower,draw=none] table[x=samples,y expr=\thisrow{snmn}-\thisrow{snmn_std}] {data/e2e_efficiency.dat};
\addplot [fill=brown!30, fill opacity=0.8] fill between[of=upper and lower];

\addplot [name path=upper,draw=none] table[x=samples,y expr=\thisrow{ours}+\thisrow{ours_std}] {data/e2e_efficiency.dat};
\addplot [name path=lower,draw=none] table[x=samples,y expr=\thisrow{ours}-\thisrow{ours_std}] {data/e2e_efficiency.dat};
\addplot [fill=blue!30, fill opacity=0.8] fill between[of=upper and lower];

\end{axis}
\end{tikzpicture}
        }
    \end{minipage}%
    \caption{A comparison of efficiencies during joint training of vision and language. We test \PROJNAME{} with 0.1 \% directly supervised labels.}
    \label{tab:e2e_comp}
\end{figure}
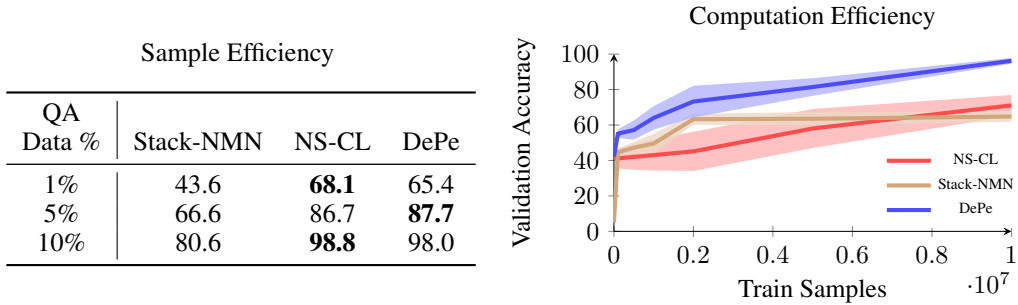

We compare the efficiencies of \PROJNAME, Stack-NMN, and NS-CL in Table \ref{tab:e2e_comp}. To test the computational efficiency of NS-CL we trained the program parser till each iteration step, fixed the parser, and then trained the vision components end-to-end given the fixed parser. 

Overall the results reflect the results reported during our component-wise testing. In terms of the computational efficiency, NS-CL \citep{NSCL} uses REINFORCE to optimize its $FP$ parser, which requires many more iterations to converge. Stack-NMN \citep{stack_nmn} is also able to optimize end-to-end, but trains the $f_I$ representation requiring more training samples. \PROJNAME{} using Stack-NMN soft programs and object-level execution from NS-CL is able to more efficiently optimize over either method alone. 

In terms of sample efficiency NS-CL and \PROJNAME{} are comparable given enough data since the models execute similarly once NS-CL language models are fine tuned. Since \PROJNAME{} is trained on some directly supervised data we also attempted directly supervising Stack-NMN and NS-CL, but saw similar performance as discussed in the model detail in Appendix \ref{apx.exp}.

\subsection{Comparative Performance}
\begin{table}[]
    \centering
    \begin{tabular}{c | c c c c | c c}
    \toprule
    Method & Vision & Language & Reasoning & Training & 100\% Data & 10\% Data\\
    \midrule
    MAC & Pixel & Text Att. & Neural & E2E & 98.9 & 67.3 \\
    TbD & Pixel & Symbolic Prog. & Neural & E2E & 99.1 & 54.2 \\
    Prob-NMN & Pixel & Symbolic Prog. & Neural & E2E & 97.1 & 88.2 \\
    Stack-NMN & Pixel & Soft Prog. & Neural & E2E & 96.3 & 86.8 \\
    LXMERT & Object & Text Att. & Neural & E2E & 97.9 & 66.5 \\
    LCGN & Object & Text Att. & Neural & E2E & 97.9 & 64.8 \\
    NGS & Object & Symbolic Prog. & Symbolic & Abduction &\textbf{100.0} & 87.3 \\
    NS-VQA & Object & Symbolic Prog. & Symbolic & RL & 99.8 & 92.1 \\
    NS-CL & Object & Symbolic Prog. & Soft Logic & E2E + RL & 99.2 & 98.8 \\
    % NGS-RL & 96.9 & 80.0 \\
    \hline
    \PROJNAME \ (0.1\%) & Object & Soft Prog. & Soft Logic & E2E & 99.0 & 98.0 \\
    \PROJNAME \ (1.0\%) & Object & Soft Prog. & Soft Logic & E2E & 99.5 & \textbf{99.3} \\
    \bottomrule
    \end{tabular}
    % \end{minipage}
    \caption{Performance of \PROJNAME{} and other VQA models using 100\% and 10\% QA pair supervision. Here we show \PROJNAME{} using 0.1\% and 1\% of the direct supervision on $O$ and $FP$.}
    \label{tab:model_comp}
\end{table}

 % The 10\% result for NGS* was taken from the closest reported result at 25\% data.

We present the accuracy comparison across different VQA models in Table \ref{tab:model_comp}. All models are able to achieve 96\%+ given the full data set, but we are more interested in the results with lower sample complexity. At 10\% data we observe that methods with continuous inference and DSL based neural modules, such \PROJNAME, NS-CL, Stack-NMN and Prob-NMN \citep{vedantam2019probabilistic} scale better. The other methods are more on the neural side, such as TbD \citep{mascharka2018transparency}, MAC \citep{hudson2018compositional}, LXMERT \citep{lxmert}, and LCGN \citep{hu2019language}, require more data to converge. Methods that discretely sample, such as NS-VQA \citep{yi2018neural} or NGS \citep{NGS}, can also achieve high accuracies given more training data, but in practice, require many iterations to converge and result in high variance results compared to continuous methods.

\section{Conclusion and Future Work}

In VQA there are different paradigms of modeling the visual, language, and reasoning representations. In addition to the final model performance, we are interested in understanding model efficiency in these complex perception and reasoning tasks. First we introduced the existing models with their corresponding vision, language, reasoning, and training components used. Then we formally defined these components and the common method representations used within each component. In order to determine which methods were the most sample and computationally efficient, we iteratively tested the vision, reasoning, and language components. These results showed that object-level, soft program, soft logic, and end-to-end training were important for efficient VQA. Following this we modified existing models to leverage all of these efficient methods in a joint manner. We showed that this model, \PROJNAME, was the most efficient while retaining state-of-the-art performance.

Based on these methods we look forward to testing \PROJNAME{} on larger real-world data sets. Since our model uses generic function programs to operate over language and vision it can be extended to different data sets with minimal modifications. Furthermore we plan to investigate how to use concepts embeddings similar to NS-CL within our memory instead of one-hot representations. We will also be interested in testing how the object-level representations work on questions involving both object and image level reasoning.

% We present our model DePe, a fully differentiable program executor over vision and text models, while leveraging an explicit set of soft function programs. Different from previous neuro-symbolic approaches that pre-generate function layout to guide program execution,  \PROJNAME{} proposes these functions at each step dynamically and executes them in a probabilistic manner. To support such an execution, we propose a heterogeneous memory to store the intermediate computation and provide the final answer. Since our approach is end-to-end differentiable, we can achieve state-of-the-art performance with fewer data. Furthermore, the sample and statistical efficiency of our model components are better than other VQA baselines.

% We look forward to reducing further our framework's data requirement, especially the pre-training requirement, by learning better inductive biases for the program parsing. Additionally, we look to apply our framework to test its performance on real image VQA benchmarks.
\newpage
\bibliography{iclr2021_conference}
\bibliographystyle{iclr2021_conference}

\newpage
\appendix
\section{Model Comparisons}
\label{apx.compare}
% We will discuss the most relative ones below: 
We explore a few comparative works that contain a variety of training strategies used for VQA. Each method type handles the vision, language, reasoning and training in different fashions.

% \begin{itemize}%[nolistsep]
\textbf{LXMERT.} Reasoning over images based on questions is carried out via Transformer like deep architecture. Such neural reasoning module can be easily interfaced with neural perception modules, and optimized end-to-end with perception module jointly. Since such model incorporate few prior structures into the model, it contains lots of parameters and requires lots of question-answer pairs in order to achieve good results.

\textbf{NMN methods.} With Neural Module Networks (NMN), the language is turned into a discrete function program over neural modules which act on pixel level attention to answer questions. The discrete function program allows reasoning steps to be executed exactly. However, such design also makes the entire model not end-to-end differentiable. One needs to use REINFORCE to optimize the model for producing the function program. 

An extension of NMN is Prob-NMN where the predicted program is supervised over a small set of samples. These ground truth programs provide a prior distribution over valid programs. This prior is used to determine future valid programs and enforce systematic program layouts.

In Stack Neural Module Network, reasoning instructions in the question are executed as a neural program over neural modules which act on pixel level attention to answer questions. This neural program execution approach produces a soft function program, where discrete reasoning structures, such as differentiable pointer and stack, are incorporated into the neural program executor. This enables Stack-NMN to maintain uncertainty over which reasoning steps are selected. 

\textbf{GNN Methods.} In Neural State Machine (NSM) and Language-Conditioned Graph Neural Networks (LCGN), images are represented as object and relations, and graph neural networks conditioned on the language feature are used as reasoning modules. Graph neural networks are structured networks, which can represent logical classifier in graphs. Such graph neural networks and deep perception architectures are end-to-end differentiable. However, the architecture is quite generic, and requires large number of question-answer pairs to train.

\textbf{NS-CL.} In Neural Symbolic Concept Learner, questions are turned into a symbolic program with soft logic function, which are then executed end-to-end on object attention produced by the vision model. The soft logic makes the reasoning step end-to-end differentiable. However, the functional programs are still discrete in structure, making the entire model not end-to-end differentiable. One needs to use REINFORCE to optimize the model for producing the function program.

\textbf{NGS.} Neural-Grammar-Symbolic performs abduction, where both the image and language are turn into discrete object by sampling from the perception model, and symbolic program is exected to generate the answers. 

In abductive learning, the symbolic reasoning step is directly interpretable, and many existing logic reasoning engine can be used. However, the model is not end-to-end differentiable. Discrete optimization is needed to generate supervision for the vision and language model to be updated. 
% \end{itemize}

\section{Differentiable End-to-end Program Executor}
\label{apx:depe}

\begin{figure}[t]
\centering
\includegraphics[width=1.0\textwidth]{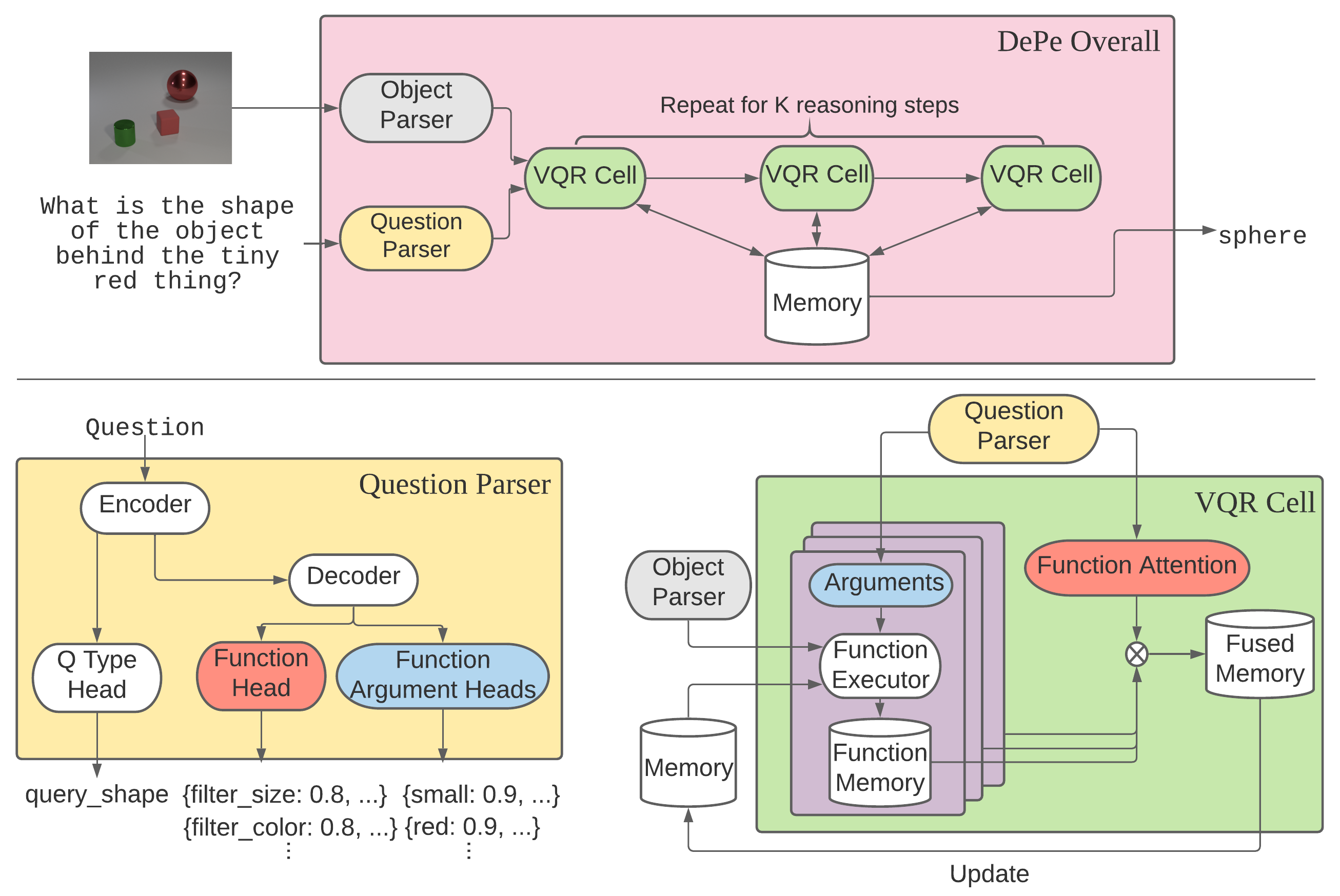}
\caption{Our model ingests the image and extracts object-centric information. The question (textual input) is used by the question parser, which first embeds the question through an LSTM encoder. This embedding is used to predict the question type, which will be used to retrieve the answer from memory at the final step. The encoded text is passed to a decoder, which at every step predicts attention over which functions to execute and each function's arguments at the current step. The objects, arguments, and any preceding memory inputs are used by each function to execute, stored in memory belonging to each function. These function memories are weighted by functional attention and fused, which updates the original memory. The decoding and the VQR Cell execution run a fixed number of times, and then the answer is extracted from the final memory.}

% \bh{The idea of maintaining a distribution of functions/programs is not illustrated very well. We should emphasize this since this is our main contribution. Maybe we need a dedicated figure for this. Now it's hard to tell the difference between our work and NSCL/stack-NMN.}

% \Karan{edit link here https://app.lucidchart.com/invitations/accept/c0da37dd-935e-4128-b26a-01ffa55a9347 export by file \-> export}
\label{fig:overview}
\end{figure}

The overall \PROJNAME{} architecture, as shown in Figure \ref{fig:overview}, consists of multiple sub-parts. The perception models that encode the vision and the question. The soft logic functions $\mathcal{F}$ closely follow the domain-specific language (DSL) provided by CLEVR. We implement them in a probabilistic manner based on the signature, detailed in Appendix \ref{apx.soft_funcitons}. The function execution results are stored in a differentiable memory that maintains a stack of these operations. As our model executes step by step, this memory is used as a buffer for the reasoning chain's probabilistic steps leading to a final answer. 

% \bh{Based on the description above, the readers cannot distinguish it with Stack-NMN. Probably should mention that we maintain a program distribution, and execute over the distribution?}

\subsection{Perception}
{\bf Object detection.} For our vision, we use object-level detection through a Mask R-CNN \citep{he2017mask} to extract the corresponding bounding boxes of the objects. We then take detected objects and pass them through a pre-trained ResNet-50 model \citep{he2016deep} to extract image features. These features are fed into models to predict object attributes from the image features and their bounding boxes' object relations.

{\bf Question parser.} The parser takes in the text input and encodes it through a Bi-LSTM to generate the final hidden state. This hidden state is used in the program parser processes where a function model predicts the attention over which soft function is to be used at that step. Additionally, these functions may contain arguments, such as $\texttt{filter\_shape[cube]}$, so for each class of functions, we have a model to predict a distribution over the possible attributes. The hidden state is also used to predict question type, which is used to select the final memory answer at the end of the execution.

For the vision and the language, all prediction models are small MLPs. We denote the set of trainable vision and text models parameters as $\theta_{vision}$, and $\theta_{text}$ respectively.

\subsection{Memory}
% \Le{Need to explain clearly how different function modules are made to take the same input and generate the same outputs, such that they can all executed by soft function program executor.}

In the CLEVR DSL, the functions have different types of inputs and outputs. Stack-NMN used neural modules to approximate these DSL functions, thus were able to make the image attention output of a consistent length. Since we are using the soft version of the DSL, we had to design a memory structure that could handle varying input and output modalities and sizes.

\textbf{Memory block.} To create a heterogeneous memory structure, we pre-assigned different modalities to different parts of a matrix. This memory matrix block $M \in \mathbb{R}^{T \times (A + C + 1)}$ is used to store the intermediate results of the function computation. 

% Unlike the memory in Stack-NMN, which just stored image attention, we have to accommodate for different output modalities for each function.

Here the first dimension $T$ is the stack dimension on which the function outputs are stored in different steps of the execution. The second $(A + C + 1)$ dimension are the rows that store the heterogeneous values. For the rows, $A$ elements in the row indicate the object set attentions $m_{det}$, or which objects the model is paying attention to at that step. The $C$ stores the categorical value outputs possible by our vision models such as $m_{color}$ or $m_{size}$. For example if there are 6 possible colors, then $m_{color} \in \mathbb{R}^6$. The final dimension $m_{num}$ is reserved for some accumulated score such as $\texttt{count}$ or a boolean bit for true or false questions. For the CLEVR specific case the object attentions $m_{det} \in \mathbb{R}^A$. The categorical values are $[m_{color}; m_{shape}; m_{texture}; m_{size}] \in \mathbb{R}^C$, and the numeric slot $m_{det} \in \mathbb{R}$.

% For example within a block of the memory row $m = M_i$, one section $m_{color}$ is attributed to the distribution of color from the previous execution. If there are 6 possible colors, then $m_{color} \in \mathbb{R}^6$. Similar sections within the categorical outputs are pre-allocated for the detections $m_{det}$, $m_{shape}$, $m_{name}$ and so on are concatenated horizontally. 

\textbf{Stack structure.} Choosing which row $t \in T$ of the memory is accessed is handled by the stack. This enables the model to accurately retrieve the previous one or two stack values as needed by the functions, instead of to predict which locations in memory to use. 

Some functions may require just looking at the previous output from the VQR Cell, such as chain-like questions over a scene graph through functions such as \texttt{filter, relate, sum}. Such functions will pop from and then push to the same memory row. There are situations where functions need multiple inputs as well, such as comparison functions \texttt{attribute\_equal, count\_equal}. These functions will thus pop two values from the stack and will only push back one. These function signatures are summarized in Table \ref{tab:funcs} in Appendix \ref{apx.soft_funcitons}.

\textbf{Stack manipulation.}   Each function will have access to the stack, which abstracts the memory management appropriately through push and pop operations. The stack memory $M$ is initialized at random and a one-hot memory pointer $p \in \mathbb{R}^T$, starting at $p_0 = 1$. Each function returns a row $m$ or a specific slice such as $m_{color}$ to be updated in memory.  

To push a row $m$ onto the stack we follow the Stack-NMN convention updating the pointer as $p = \text{1d\_conv}(p, [0, 0, 1])$ and the memory row 
$M_i = M_i \cdot (1 - p_i) + m \cdot p_i$. Here the convolution is just moving the one hot pointer left by one in a differentiable manner. Then only the memory row that contains the one hot pointer is updated with the row $m$. Similarly to pop a value we perform the following retrieve the row $m = \sum_{t=1}^T p_i \cdot M_i$ and push back the pointer right by $p = \text{1d\_conv}(p, [1, 0, 0])$.

% \[m = \sum_{t=1}^T p_i \cdot M_i\]
% \[p = \text{1d\_conv}(p, [1, 0, 0])\]

% which retrieves the row of the active pointer and shifts the pointer back right.

\subsection{VQR Cell}
\begin{figure}[t]
\centering
\includegraphics[width=1.0\textwidth]{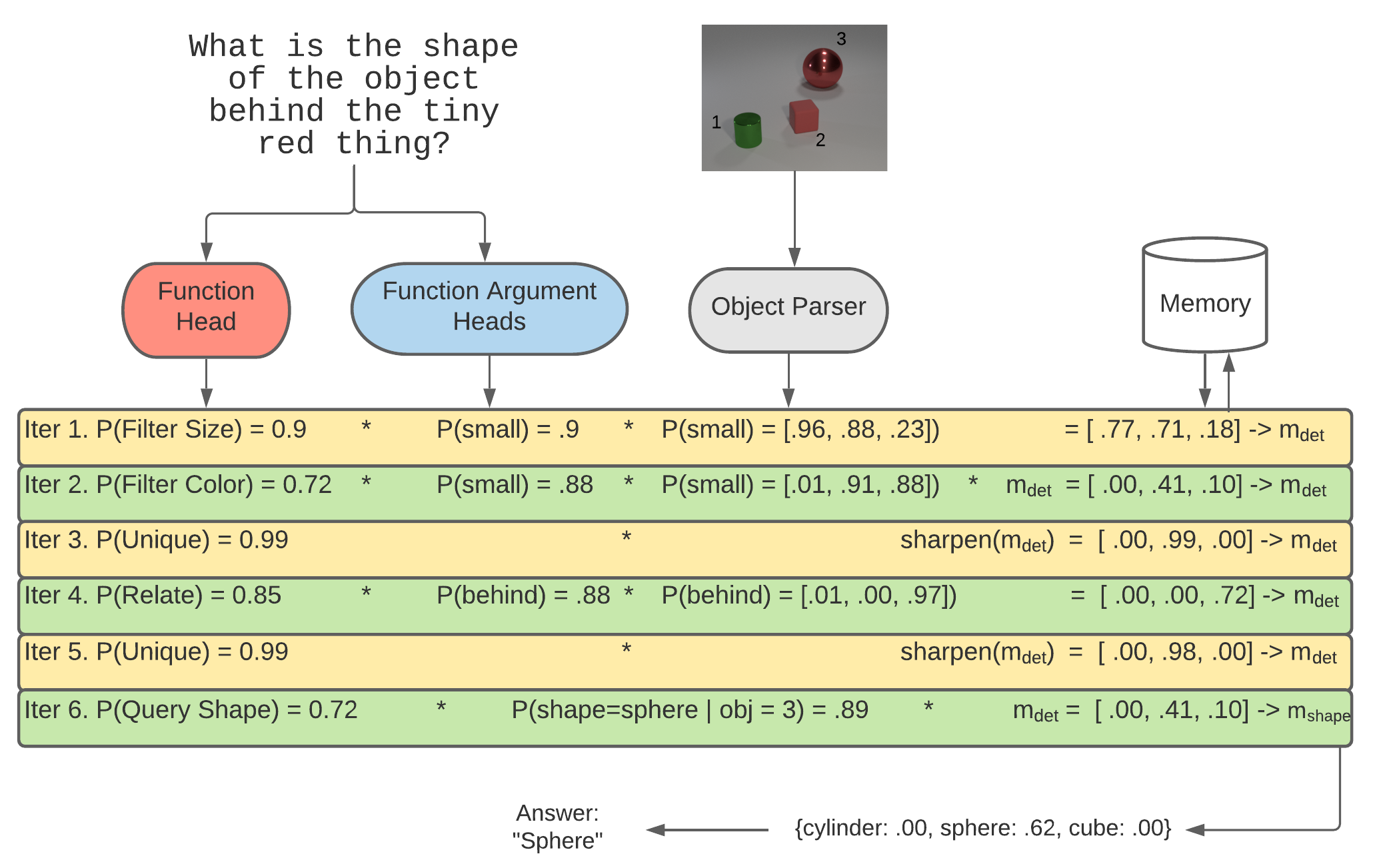}
\caption{Here is a sample execution for the vision and text in DePe. For the simplicity of the diagram, we don't include all the probability traces and just show the ones corresponding to the max probability. For the vision, the predictions correspond to the objects detected in the scene (marked with 1, 2, 3 in the picture). For each execution, we store the multiplication in the corresponding memory row, and retrieve the final memory to answer the question.}
\label{fig:exec}
% \bh{The table is somehow misleading since we only compute using the trace with the max probability. I feel like we need to have an example where multiple functions are included in the computation. Could you shorten the question and include at least two functions each iter?}

\end{figure}

The visual question reasoning (VQR) Cell is responsible for the step by step executions over the vision and the text to get to the answer. Compared to previous methods that executed these DSL functional programs in a discrete sequence, we generate probabilistic output over all the functions $\mathcal{F}$ at each step. With this approach, we can propagate the uncertainties of the object detections \textit{and} the question parser for end-to-end learning. An example of this execution is visible in Figure \ref{fig:exec}.

% The execution for each soft function is determined by three possible types of inputs: (1) Memory inputs $m$ from the previous cell iteration. (2) Object detection predictions, and (3) function arguments. 
% \begin{itemize}
%     \item Memory inputs $m$ pertaining to the information stored by a function in the previous VQR Cell iteration.
%     \item Object detection predictions such as $P_{rel}, P_{attr}$ representing the object relation and attribute predictions.
%     \item Argument predictions for the function $p_{arg}$.
% \end{itemize}

\textbf{Soft function execution.} Each function has different return values and signatures. Since functions can have different input and output requirements, they require to be at different positions of the stack. Therefore each function is given its copy of the memory stack to operate over.

\begin{minipage}[t]{8cm}
Once the memory outputs for each function are computed, they have to be weighted by which function our model is most likely needed for this step. This is done by using the function attention weights $w$ computed by the question parser. These weights scale each function's memory and the pointer and then computes a weighted sum, which is then used to update the global memory $M$ and pointer $p$, as shown in Algorithm \ref{alg:cell}. This global memory is then copied for all the function memories in the next Cell iteration. The cell executes for a fixed number of iterations and is set to a number such that it can cover most ground truth program lengths.
\end{minipage}%
\hfil %
\begin{minipage}[t]{5cm}
\begin{algorithm}[H]
\SetAlgoLined
% \KwResult{Expected Memory $M_{i+1}$}
%  \For{$f \in \mathcal{F}$}{
 \For{$j \in [1, |\mathcal{F}|]$}{
    $M_j, p_j = \mathcal{F}_j(M, p)$
 }
 $M_{ave} = \sum_{j=1}^{|\mathcal{F}|} w_j \cdot M_j$ \;
 $p = \text{softmax}(\sum_{j=1}^{|\mathcal{F}|} w_j \cdot p_j)$ \;
 $M = M \cdot (1 - p) + M_{ave} \cdot p$ \;
%  M.push($m_{ave}$)
\caption{VQR Cell iteration}
\label{alg:cell}
\end{algorithm}
\end{minipage}%
% \textbf{Memory Aggregation.} To accommodate each modular function execution, which may require different number of push or pop operations, each function runs on a copy of the memory. Then the memory is collated based on the module weights $w_i$ as indicated in the VQR cell execution in \ref{alg:cell}.

\textbf{Final answer.} After the last iteration of VQR Cell execution, the final answer is chosen by the memory corresponding to most probable question type. For example if we predict the question is asking about the color, then we return argmax($m_{color}$). If we predict the question type is about count, then we return $m_{num}$.
Since our memory structure is heterogeneous, taking argmax over the entire row $m$ may lead to an incorrect label due to parts of the memory storing values at different ranges, such as probabilities $\in [0, 1]$ or counts $\in \mathbb{Z}^+$.

% Intuitively, we are trying to understand which probabilistic combination of these function modules leads to the final answer. One may also make a parallel comparison of this optimization process to differentiable neural architecture searches, wherein those problems they search for an arbitrary architecture composed of convolution or linear layers. In our case, we want to define architectures composed of DSL functions, but conditioned on the question and image. Since these soft function weights are continuous from the question parser, they can be optimized jointly with the perception model.

% Maybe move the "Case Study" section here to show a sample execution

% $\Theta_T = \{\theta_{lstm}, \theta_{weight}, \theta_{arg}^*\}$, $\Theta_V = \{\theta_{attr}^*, \theta_{rel}\}$ respectively.

% In order to merge the differentiable models with the soft logic functions we had to cr, we had to address the following challenges:

% \begin{enumerate}
%     \item Parameterize a question program functions and their attributes in a continuous manner
%     \item Define a heterogenous memory structure to track specific object attributes and intermediate computation
%     \item Infer and train on the heterogeneous memory stack and program functions end-to-end
% \end{enumerate}

\subsection{Optimization}
The objective is to minimize the overall loss based on the predictions retrieved from memory. This involves a multi-task loss as we are predicting categorical and numeric answers to questions to optimize over $\theta_{text}, \theta_{vision}$. This is done by minimizing the cross entropy loss for categorical answers and a mean squared error loss for numeric ones: $\mathcal{L} = \mathcal{L}_{QA\_XEnt} + \mathcal{L}_{QA\_MSE}$.

This optimization, particularly for the text has posed to be a difficult problem from scratch as there are many candidate functions at each step in the function program. Furthermore spurious programs can be proposed in early stages of training, corresponding to the shortest programs that answer the question, but don't align with the ground truth semantics.

To address this, models such as NS-CL, employ structural constraints when training their text parser. They employ a fixed number of concepts tokens used to parse the question and discrete templates for the parser to follow. In addition they leverage curriculum learning and switch between optimizing the parser and vision models to support both the discrete and continuous training methods.

Other models, such as NS-VQA, train on a small portion of the supervised labeled data. In this manner no such restrictions on the concepts, templates, or curriculum learning, but require labeling of ground truth programs. In this work we explore the pre-trained route. We pre-train our models with a small percentages (0.1-1\%) of visual and textual data. When we use X\% pre-training data, we sample X\% of the training questions. Then we directly supervise our question parser and vision models on those QA programs and their corresponding vision scenes.

Additionally, we found it useful to add these direct supervision losses when training end-to-end with corresponding weights $\alpha$ and $\beta$. We formulate this as a regularization method that prevents the models from diverging to spurious predictions when the training signal is only coarse QA labels. 
\[\mathcal{L} = \mathcal{L}_{QA\_XEnt} + \mathcal{L}_{QA\_MSE} + \alpha \mathcal{L}_{Obj\_XEnt} + \beta \mathcal{L}_{Text\_XEnt}\]

The full definitions and learning strategies for the losses are available in Appendix \ref{apx.optim}.

\section{Optimization Details}
\label{apx.optim}

If given and Image $I$ and a question $Q$ our model makes a prediction as follows. We first make predictions of the objects and text components as:

% \[\hat{obj} = \text{ObjectParsing}(I; \Theta_V)\]
% \[\hat{text} = \text{QuestionParsing}(Q; \Theta_L)\]
% \[\hat{y} = \text{\PROJNAME}(\hat{obj}, \hat{text}; \Theta_T, \Theta_V)\]
\[\hat{y} = \text{\PROJNAME}(I, Q; \theta_{text}, \theta_{vision})\]

We look to minimize the following loss $\mathcal{L}$.

\[\mathcal{L}_{QA\_XEnt} = -\frac{1}{N} \sum_i^N \sum_j^C \mathbb{I}_{y_i \in cat} \cdot y_{ij} \text{ log } \hat{y}_j\]

\[\mathcal{L}_{QA\_MSE} = \frac{1}{N}\sum_i^N \mathbb{I}_{y_i \in num} \cdot ||y_{i} - \hat{y}_{num}||_2\]

\[\mathcal{L} = \mathcal{L}_{QA\_XEnt} + \mathcal{L}_{QA\_MSE}\]

We note that for stable convergence we include the pre-training data to our loss function. These are the cross entropy loss for the object attribute and relation predictions $\mathcal{L}_{Obj\_XEnt}$. These are also the cross entropy loss for the question parser predictions of the functions at each step $\mathcal{L}_{Text\_XEnt}$. This gives us our final loss function: 

\[\mathcal{L} = \mathcal{L}_{QA\_XEnt} + \mathcal{L}_{QA\_MSE} + \alpha \mathcal{L}_{Obj\_XEnt} + \beta \mathcal{L}_{Text\_XEnt}\]

We include two scalar terms for the pre-training losses to balance the models behavior to find local minima within the QA losses and focusing too much on optimizing the smaller number of pre-training samples. Setting these hyperparameters is more prelevant when the ratio of pre-training to QA samples is low, since the pre-training samples provide a weak signal, but should be respected. We find that setting $\alpha = \beta = \sqrt{\text{\% of QA data}}$, works well in practice to balance this ratio. The gains of setting these hyperparmeters can be seen in Table \ref{tab:loss_weights}.

% When training end-to-end, we observe a large performance difference by weighting the pre-training losses $\alpha$ and $\beta$. With $\alpha = \beta = 1$ we observe that the model attempts to balance between the QA loss and the pre-training losses. This still leads to spurious predictions when there are more QA labels than pre-training ones. Therefore it is important to weight these hyperparameters to make sure that the model is consistent with the ground truth vision and language labels, as seen in Table \ref{tab:loss_weights}.

\begin{table}[h]
    \centering
    \begin{tabular}{c c c c c c c c c c c}
        \toprule
        % & & & pre-training\% & & \\
        % \cmidrule{3-5}
        & & \multicolumn{3}{c}{0.1\% pre-train} & \multicolumn{3}{c}{0.5\% pre-train} & \multicolumn{3}{c}{1\% pre-train} \\
        \% QA Data & $\alpha=\beta$ & base & weight & $\Delta$ & base & weight & $\Delta$ & base & weight & $\Delta$ \\
        \midrule
        5\% & 3 & 87.7 & 97.9 & 10.2 & 97.1 & 99.1 & 2.0 & 98.8 & 99.2 & 0.4 \\
        10\% & 6 & 92.3 & 98.0 & 5.7 & 96.6 & 99.2 & 2.6 & 99.1 & 99.3 & 0.2 \\
        100\% & 10 & 93.1 & 99.0 & 5.9 & 97.7 & 99.4 & 1.7 & 99.4 & 99.5 & 0.1\\
        \bottomrule
    \end{tabular}
    \caption{Absolute validation accuracy gains by adding pre-training loss weights. The gains were computed against the baseline method of $\alpha = \beta = 1$.}
    \label{tab:loss_weights}
\end{table}

We use Adam to optimize with $lr = 1e^{-4}, \beta_1 = 0.9,\beta_2 = 0.999, \epsilon = 1e^{-8}$ and set our batch size to 1024.

\section{Experiments}
\label{apx.exp}

Here we outline the details of the training procedure for the end-to-end training and efficiency experiments. We follow the default training procedure for the relevant work if it is not modified in this section.

\subsection{LXMERT}
For LXMERT we were interested in exploring if the cross modal architecture could implicitly encode our attention based reasoning process, as conveyed by the recent Transformer literature. There were a couple of approaches that we tried during our tests.

The initial tests and the results we reported were the based on the original architecture proposed in the paper as well as using the pre-trained weights the authors provided. This used 9, 5, and 5 layers of the language encoder $N_L$, the cross modality encoder, and the region of interest (RoI) object encoder $N_R$ respectively.

We also test the initializing the RoI encoder from scratch since the CLEVR features are more simpler than the ones used in pre-training on real images. Additionally we tested LXMERT with fewer layers, $N_L = 4, N_X = 2, N_R = 2$, which we report in Table \ref{tab:lxmert}.

\begin{table}[!h]
    \centering
    \begin{tabular}{c c c c}
    \toprule
         \% QA Data & LXMERT Pre-trained &  LXMERT Scratch & LXMERT 422 \\
         \midrule
         100 & 97.9 & \textbf{98.0} & 95.6 \\
         50 & \textbf{95.3} & 93.5 & 90.4 \\ 
         25 & \textbf{89.2} & 69.27 & 70.71 \\
         10 & \textbf{66.5} & 50.55 & 52.72 \\
         \bottomrule
    \end{tabular}
    \caption{LXMERT CLEVR performance on different subsets of data and configurations.}
    \label{tab:lxmert}
\end{table}

\subsection{Stack-NMN}
When experimenting with Stack-NMN we first trained the entire model on 100\% of the CLEVR data using the ground truth layout and save it. For the visual representation experiments we take the trained model and freeze the LSTM weights while resetting the vision weights. For the language representation we freeze the CNN and NMN weights without the expert layout.

When training Stack-NMN for direct supervision, we train the model with the ground truth layout but only on 0.1\% of the  data. Then we take load this model and ran it on the corresponding 1, 5, and 10\% QA data without the expert layout. With only 0.1\%, we didn't see any sample of computational performance differences on end-to-end training.

We used the same settings as the original paper and trained all variations for 200k iterations.

\subsection{NS-CL}

When testing NS-CL end-to-end we require iterating between optimizing the language program parser and training the vision concept models, done in a REINFORCE and end-to-end fashion respectively. When conducting the computational complexity experiments we had to train and test the program parser and vision models at fixed number of training samples. To do this we divide up the total number of training samples uniformly into 10 curriculum lessons used in the original NS-CL. We then tune the program parser with the data belonging to the cumulative curriculum up to those number of training samples. Given this program parser we tune the vision models end-to-end over the same number of training samples. This way we have a better picture of NS-CL's behavior as a function of the training iterations.

To make NS-CL more comparative to \PROJNAME, we also use the 0.1\% direct supervision data improve the NS-CL initial performance. We similarly train the program parser on the $(Q, FP)$ pairs, thus start off at similar accuracies. Unlike \PROJNAME, NS-CL uses similarities between text and vision concepts, soit was unclear how to directly supervise the vision model given the object data. Similar to Stack-NMN we test a burn-in strategy where we train the entire model end-to-end with the QA Data from the corresponding 0.1\% direct supervision then train using the 10\% QA data for 150 epochs. Here we observed no significant gains from just start off with the supervised vision parser.

In the reasoning representation experiments we test NS-CL with discrete reasoning. Here we took the original soft logic $\mathcal{F}$ defined in the NS-CL paper and replaced them with symbolic logic as used in NS-VQA. Then we used REINFORCE to optimize when training on QA data in NS-RL. In NS-AB we use abduction to find the most likely change in the object concept prediction to make the resulting reasoning provide the correct answer. We discuss this in the next section. 

\subsection{NS-AB Abduction}
\label{apx.exp.ngs}

% \subsubsection{Overview}
% Describe our method of carrying out the abduction based execution, since at the writing of this paper, the code to test NGS on CLEVR has not been released yet. 
Abduction  involves  determining  which detection $m_{det}$ changes are required to correctly answer the question given a fixed discrete function program.

At a high level these changes are done by making the most likely change that makes the output answer right. In NGS they use a tree based formulation and in our implementation we retain our stack implementation. Our stack implementation lets us test abduction proposals in a greedy efficient manner, but leads to a smaller search of the detection space leading to the ground truth answer instead of a spurious one. We describe the details of our implementation below.

\subsubsection{Policy Definition}
The function program $Q=\{f_i\}^T, f_i\in \mathcal{F}$ tells us to execute a sequence of $T$ operations or functions $f_i$. If there are multiple reasoning branches that are aggregated, the execution trace follows a DAG dependency structure. We use a Markov decision process (MDP) formulation for the execution process:   
\begin{itemize}
    \item A policy $m_{det}^t\sim \pi_{\theta}(m_{det}^t | m_*^t,f_t,I)$ with vision models parameters $\theta_{vision}$ will take the operation execution from the previous stage $m_*^t$, the operator indicator $f_t$ and image $I$, and output an action $a_t$. Memory from the previous stage $m_*^t$ could be any memory slice such as $m_{det}, m_{attr}, m_{num}$. This action $m_{det}^t$ corresponds to the object selections made for the current operation from the input image. 
    \item The action will cause the state $m_*^t$ to be updated to $m_*^{t+1}$, and the transition is described by $P(m_*^{t+1}|m_*^t,m_{det}^t,f_t)$. This is the operation execution for the current stage. This is carried by logic function $f_t$ as described in Table \ref{tab:funcs}.
\end{itemize}
After this sequence of operations, we obtain the a reward $R(A, m_*^{t+1})$ by comparing the last selected state with the answer A, and tell us whether the answer is correct or not. If the answer is correct $R=1$, and otheriwse 0. Then the expected correctness of answer given the uncertainty in $\pi_{\theta}$ and $P$ is 
\begin{align}
    &\mathbb{E}[R(A,m_*^{t+1})],~\text{where}~\\
    &m_*^{t+1}\sim P(m_*^{t+1}|m_*^t,m_{det}^t,f_t),~a_t\sim \pi_{\theta}(m_{det}^t|m_*^t,f_t,I),~m_*^0=\emptyset,~\forall t=1,\ldots,T
\end{align}

% and the value $r$ from previous transformation as input and output a different set of objects: 
% \begin{align}
%     (S_{i+1}, r_{i+1}) = T_i[(S_i, r_i)] 
% \end{align}
% And the probability of producing the output is according to probability:
% \begin{align}
%     (S_{i+1}, r_{i+1}) \sim P[(S_{i+1}, r_{i+1})]
% \end{align}
% In fact, only those transformations involving object selection/attribute prediction from the image will have a probability attached to the output. Other numerical or deterministic operations do not need a probabilistic model. \Le{notation need to be made better.}
\subsubsection{Optimization}

Given $m$ data points expressed as image $I$, questions $Q$ and answer $A$ triplets: $(I_1,Q_1,A_1),\ldots,(I_m,Q_m,A_m)$, the learning problem can be expressed as: 
\begin{align}
    \pi_{\theta}^* = \argmax_{\pi_{\theta}} ~\sum_{i=1}^m \mathbb{E}[R(A_i,m_*^{i, t+1})]
\end{align}

{\bf Label abduction.} If for a particular data point $(I,Q,A)$ the answer according to the model is incorrect, one can try to find corrections by making minimal corrections $c_i$ to the model detections as follows:
\begin{align}
    c_1^*,\ldots,c_T^* = \text{argmin}_{c_1,\ldots,c_T}~& D(m_{det}^1,\ldots,m_{det}^T \| c_1,\ldots,c_T)  \\
    \text{s.t.}~& R(A,m_*^{t+1}) = 1 ~~\text{makes the answer right.} \\
    & m_*^{t+1}\sim P(m_*^{t+1}|m_*^t,c_t,f_t), \forall t = 1,\ldots,T  
\end{align}
where $D(\|)$ is some distance measure between the corrections $c_i$ and the model predictions $m_{det}^i$. For instance, $D(\|)$ can be the negative likelihood of the correction under our model $\pi_{\theta}$, or the edit distance. 

Intuitively, we want to attempt the most likely corrections iteratively to find a solution by minimizing $D(\|)$. Sampling all possible corrections at once, can cause right answers through spurious label changes, which we want to mitigate.

{\bf Sampling methods.} Due to the compositional nature of the reasoning tasks, we attempt to optimize $c_i^*$ in a greedy fashion at each step of the program from $i=1$ to $i=T$ instead of jointly from $i \subseteq \{1, \ldots, T\}$. 

This better enforces the consistency constraint when a single $c_i^*$ update leads to valid program executions. Valid executions mean that the predicted or abduced labels $c_i^*$ lead to a final answer, whether right or wrong. This is opposed to making conflicting changes in all $c_i^*, ~ c_j^*$ where $i \neq j$, leading to a failed program execution due to the manual abduced changes. If this greedy approach fails to find an answer, we fall back on exhaustively sampling $m_{det}^i\sim \pi_{\theta}$ at all program levels for a fixed number of iterations.

\section{Modular Soft-Logic Functions}
\label{apx.soft_funcitons}

The descriptions of the variables and constants used to describe memory components are listed in Table \ref{tbl:vars}. The functions used are in Table \ref{tab:funcs}. For notation simplicity, the function arguments are assumed to be popped from that function's memory copy or predicted from the text or vision models.

\renewcommand{\arraystretch}{1.5}
\begin{table}[h!]
\centering
\begin{tabular}{c c c c}
    \hline
    Name & Module & Description & Values \\
    \hline
    D & Hyperparameter & Number of objects & $\mathbb{Z}^+$ \\
    \hline
    $\gamma$ & Hyperparameter & Shift value & $\mathbb{R}$ \\
    \hline
    $\tau$ & Hyperparameter & Scalar value & $\mathbb{R}$ \\
    \hline
    $m_{det}$ & Memory & Gates determining the active object detections & $\{[0, 1]\}^D$ \\
    \hline
    $m_{num}$ & Memory & Storage for numerical and boolean operations & $\mathbb{R}$ \\
    \hline
    $m_{attr}$ & Memory & Probability distribution over that attribute & $\mathbb{R}^{|attr|}$ \\
    \hline
    $p_{arg}$ & Text & Function argument probability distribution & $\mathbb{R}^{|attr|}, \mathbb{R}^{|rel|}$ \\
    \hline
    $P_{rel}$ & Vision & Object pairwise relation predictions & $\mathbb{R}^{D \times D \times |rel|}$ \\
    \hline
    $P_{attr}$ & Vision & Image attribute prediction per object & $\mathbb{R}^{D \times |attr|}$ \\
    \hline
\end{tabular}
\caption{Description of functional arguments and constants.}
\label{tbl:vars}
\end{table}
\begin{table}[h!]
\centering
\begin{tabular}{c c}
    \toprule
    Signature & Implementation \\
    \midrule
    scene() & $m_{det} := 1$ \\
    \midrule
    % unique($m_{det}$) & $m_{det} := \text{softmax}(m_{det} / \tau)$ \\ 
    unique($m_{det}$) & $m_{det} := (\frac{m_{det}}{1 - m_{det}})/\text{sum}(\frac{m_{det}}{1 - m_{det}})$ \\ 
    \midrule
    count($m_{det}$) & $m_{num} := \text{sum}(m_{det})$ \\
    \midrule
    exist($m_{det}$) & $m_{num} := \text{max}(m_{det})$ \\
    \midrule
    intersect($m_{det}^1, m_{det}^2$) & $m_{det} := \text{min}(m_{det}^1, m_{det}^2)$ \\
    \midrule
    union($m_{det}^1, m_{det}^2$) & $m_{det} := \text{max}(m_{det}^1, m_{det}^2)$ \\
    \midrule
    equal\_integer($m_{num}^1, m_{num}^2$) & $m_{num} := \text{sigmoid}((\tau - |m_{num}^1 - m_{num}^2|)/ (\gamma \cdot \tau))$ \\
    \midrule
    greater\_than($m_{num}^1, m_{num}^2$) & $m_{num} := \text{sigmoid}((m_{num}^1 - m_{num}^2 - \gamma)/ \tau)$ \\
    \midrule
    less\_than($m_{num}^1, m_{num}^2$) & $m_{num} := \text{sigmoid}((m_{num}^2 - m_{num}^1 - \gamma)/ \tau)$ \\
    \midrule
    % relate($m_{det}$, $P_{rel}$, $p_{arg}$) & $m_{det,i} := \text{min}(m_{det, i}.\text{tile}(|rel|), P_{rel, i})^T p_{arg}$ \\
    % relate($m_{det}$, $P_{rel}$, $p_{arg}$) & $m_{det} := \sum_{i=1}^D m_{det, i} \cdot ((P_{rel})(p_{arg}))_i$ \\
    relate($m_{det}$, $P_{rel}$, $p_{arg}$) & $m_{det} := (m_{det})^\top(P_{rel})(p_{arg})$ \\
    \midrule
    % filter\_attr($m_{det}$, $P_{attr}$, $p_{arg}$) & $m_{det,i} := \text{min}(m_{det, i}.\text{tile}(|attr|), P_{attr, i})^T p_{arg}$ \\
    filter\_attr($m_{det}$, $P_{attr}$, $p_{arg}$) & $m_{det} := \text{min}(m_{det}, (P_{attr})(p_{arg}))$ \\
    \midrule
    % query\_attr($m_{det}$, $P_{attr}$, $p_{arg}$) & $m_{attr,i} := \text{min}(p_{arg}, (P_{attr, i})(m_{det, i}.\text{tile}(|attr|))^T) $ \\
    query\_attr($m_{det}$, $P_{attr}$, $p_{arg}$) & $m_{attr} := \text{min}(p_{arg}, (P_{attr})^\top(m_{det})) $ \\
    \midrule
    % same\_attr($m_{det}$, $P_{attr}$) & $m_{det} := \sum_{i=1}^D m_{det, i} \cdot ((P_{attr})(P_{attr})^T - I)_i$ \\
    same\_attr($m_{det}$, $P_{attr}$) & $m_{det} := ((P_{attr})(P_{attr})^\top \odot (1 - I))(m_{det})$ \\
    \midrule
    equal\_attr($m_{attr}^1, m_{attr}^2$) & $m_{num} := (m_{attr}^1)^\top(m_{attr}^2)$ \\
    \bottomrule

\end{tabular}
\caption{Implementation details for each modular function. $m_{det}, m_{num}, m_{attr}$ correspond to different parts in the memory representing attentional masks, numerical results, and attributes. $p_{arg}$ is the distribution of functional arguments produced by the question parser, while $P_{rel}$ and $P_{attr}$ are relation and attribute predictions given by the perception module. Hyper-parameters $D=50$, $\tau = 0.25$ and $\gamma = 0.5$ and attribute functions are split further by each $attr \in \{shape, color, material, size\}$.}
\label{tab:funcs}
\end{table}

% Additional details are laid out in Appendix \ref{appendix:abduction}.

% $\hat{obj} = $ $\hat{y} = \text{\PROJNAME}(I, Q; \Theta_T, \Theta_V)_j$, we minimize the following loss $\mathcal{L} = \mathcal{L}_{QA\_XEnt} + \mathcal{L}_{QA\_MSE}$. The definitions for each loss function used are as follows:

\end{document}